# A Converged Algorithm for Tikhonov Regularized Nonnegative Matrix Factorization with Automatic Regularization Parameters Determination

Andri Mirzal

*Abstract*—We present a converged algorithm for Tikhonov regularized nonnegative matrix factorization (NMF). We specially choose this regularization because it is known that Tikhonov regularized least square (LS) is the more preferable form in solving linear inverse problems than the conventional LS. Because an NMF problem can be decomposed into LS subproblems, it can be expected that Tikhonov regularized NMF will be the more appropriate approach in solving NMF problems. The algorithm is derived using additive update rules which have been shown to have convergence guarantee. We equip the algorithm with a mechanism to automatically determine the regularization parameters based on the L-curve, a well-known concept in the inverse problems community, but is rather unknown in the NMF research. The introduction of this algorithm thus solves two inherent problems in Tikhonov regularized NMF algorithm research, i.e., convergence guarantee and regularization parameters determination.

## I. INTRODUCTION

**T**HE nonnegative matrix factorization (NMF) is a technique that decomposes a nonnegative matrix into a pair of other nonnegative matrices. Given a nonnegative matrix $\mathbf{A}$, the NMF seeks to find two nonnegative matrices $\mathbf{B}$ and $\mathbf{C}$ such that:

$$\mathbf{A} \approx \mathbf{BC}, \tag{1}$$

where $\mathbf{A} \in \mathbb{R}_+^{M \times N} = [\mathbf{a}_1, \ldots, \mathbf{a}_N]$, $\mathbf{B} \in \mathbb{R}_+^{M \times R} = [\mathbf{b}_1, \ldots, \mathbf{b}_R]$, $\mathbf{C} \in \mathbb{R}_+^{R \times N} = [\mathbf{c}_1, \ldots, \mathbf{c}_N]$, $R$ denotes the number of factors which usually is chosen so that $R \ll \min(M, N)$, and $\mathbb{R}_+^{M \times N}$ denotes $M$ by $N$ matrix with nonnegative entries. The conventional method in computing $\mathbf{B}$ and $\mathbf{C}$ is by minimizing the distance between $\mathbf{A}$ and $\mathbf{BC}$ in Frobenius norm,

$$\min_{\mathbf{B}, \mathbf{C}} J(\mathbf{B}, \mathbf{C}) = \frac{1}{2} \|\mathbf{A} - \mathbf{BC}\|_F^2 \text{ s.t. } \mathbf{B} \geq \mathbf{0}, \mathbf{C} \geq \mathbf{0}. \tag{2}$$

In addition, other distance like Kullback-Leibler divergence [1], [2] and Csiszárs $\varphi$-divergence [3] can also be used.

Previously, the NMF was studied under the term positive matrix factorization by Paatero et al. [4], [5]. The popularity of the NMF is due to the work of Lee and Seung [6] in which they introduced a simple yet powerful NMF algorithm, and then show its applicability in image processing and text analysis.

A. Mirzal is with the Faculty of Computer Science and Information Systems, University of Technology Malaysia, 81310 Johor Bahru, Malaysia e-mail: andrimirzal@utm.my

In addition, the algorithm also produces sparser factors (thus requires less storage) [6]–[8] and can give more intuitive results compared to other subspace approximation techniques like Principal Component Analysis (PCA) and Independent Component Analysis (ICA) [6], [7], [9]. Due to these reasons, many works explored the possibility of applying the NMF in some problem domains, e.g., document clustering [10], [11], spectral analysis [12], image processing [7], [8], blind source separation [14], and cancer detection [15]–[17], and showed that the NMF can give better results.

Recently, various works have been conducted to extend the standard NMF formulation (eq. 2) to also include auxiliary constraints such as sparseness [7], [8], [16], [17], smoothness [10], [12], [13], and orthogonality [19]–[24]. These constraints are usually formulated based on inherent properties of the data and prior knowledge about the applications, so that computed solutions can be directed to have desired characteristics. Algorithms for solving the problems are mainly based on multiplicative update rules algorithms. This is due to the convenience of deriving algorithm directly from corresponding objective function. However, as multiplicative update rules based NMF algorithms do not have convergence guarantee [9], [18], [19], [25], the development of converged algorithms for various NMF objectives with auxiliary constraints is an open research problem. Note that even though some alternating nonnegativity-constrained least square based NMF algorithms (e.g., projected gradient methods [25], [27], projected quasi-Newton method [28], active set method [29], and block principal pivoting method [30]) do have convergence guarantee, due to the complexity of the algorithms, it's not always clear how to incorporate those auxiliary constraints into the algorithms.

In this work, we propose a converged algorithm for Tikhonov regularized NMF using additive update rules. The additive update rules based algorithm for standard NMF first appeared in the work of Lee & Seung [1], but the convergence proof was given by Lin in ref. [18]. As in the multiplicative update rules version, the additive update rules based algorithm can also be derived directly from corresponding NMF objective, thus providing a convenient way in deriving converged algorithms for various NMF objectives.

We choose Tikhonov regularization as the auxiliary constraint because this constraint has been used in many applications, e.g., text mining [10], spectral data analysis [12], [13], microarray data analysis [29], and cancer class discovery [16] (in some works, sparseness is enforced using $L_2$ norm on the solution, i.e., the Tikhonov regularization, instead of $L_1$

norm—the more appropriate constraint for enforcing sparseness [31]), and showed that it can offer better results compared to the results of standard NMF. This constraint also can reduce influence of noise and other uncertainties in the data [9], [12], [13]. In addition, from the inverse problem study, it is known that Tikhonov regularized least square (LS) is the more preferable form in solving inverse problems because solutions of the conventional LS tend to be unstable and dominated by data and rounding errors [32]–[34]. Moreover, in the presence of noise, frequently the conventional LS solutions are rather undesirable as it leads to amplification of noise in the direction of singular vectors with small singular values [35]. Since LS is the building block of the NMF,

$$\|\mathbf{A} - \mathbf{BC}\|_F^2 = \sum_{n=1}^{N} \|\mathbf{a}_n - \mathbf{Bc}_n\|_F^2, \tag{3}$$

then it can be expected that Tikhonov regularized NMF will be the more appropriate approach to solving NMF problem in eq. 1.

Introducing Tikhonov regularization into the NMF brings the issue of how to properly determine the regularization parameters. From the inverse problems study it is known that the effectiveness of regularization methods depends strongly on the parameters; too much regularization creates a loss of information, and too little regularization leads to a solution that is dominated by noise components and has similar problems as in the unregularized solutions. There are two methods that are usually be used in determining an appropriate value for the regularization parameter: the L-curve and Morozov discrepancy principles. In this paper we will utilize the L-curve since the Morozov discrepancy principles require knowledge of the error level in the data which is often inaccessible [36].

The L-curve is a graphical tool that displays the trade-off between approximation error and solution size as the regularization parameter varies. In this curve, the proper value for the regularization parameter is the value associated with corner of the curve where both solution and approximation error have minimum norms [32], [34]. There are some methods proposed to find this L-corner, e.g., [32], [34], [37], [38]. We will use a method proposed by Oraintara et al. in ref. [34] in which they defined L-corner to be the point of tangency between L-curve with positive curvature and a straight line of negative slope. Fig. 1 shows such condition for L-corner. We choose this method because it has convergence guarantee and is relatively faster to compute than the standard method; the maximum curvature approach [32].

## II. TIKHONOV REGULARIZED NMF

Tikhonov regularization is a method for regularizing solutions of linear inverse problems in order to enhance stability of the solutions and reduce the observational errors. The method was developed independently by Phillips [39] and Tikhonov [40]. In statistic it is also known as ridge regression. Because Tikhonov regularization can smooth the solutions, it is often used as a smoothness constraint.

Given a linear inverse problem,

$$\mathbf{y} = \mathbf{Ax} \tag{4}$$

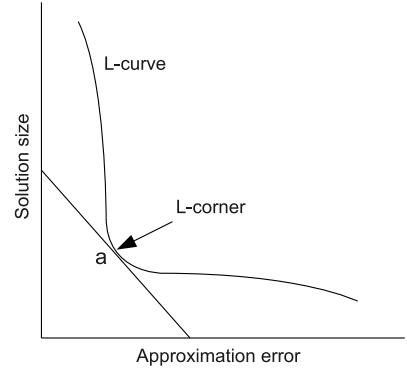

Fig. 1. The generic L-curve with positive curvature; **a** is the L-corner.

where $\mathbf{x} \in \mathbb{R}^M$ denotes unknown vector to be estimated, $\mathbf{y} \in \mathbb{R}^N$ denotes observation data, and $\mathbf{A} \in \mathbb{R}^{N \times M}$ denotes distortion matrix. The classical approach is to use standard LS approach to estimate $\mathbf{x}$,

$$\mathbf{x} = \arg\min_{\mathbf{x}} \|\mathbf{y} - \mathbf{Ax}\|_F^2. \tag{5}$$

To improve the solution, usually Tikhonov regularized LS is used instead [32]:

$$\mathbf{x}_\lambda = \arg\min_{\mathbf{x}} \|\mathbf{y} - \mathbf{Ax}\|_F^2 + \lambda \|\mathbf{x}\|_F^2 \tag{6}$$

where $\lambda$ denotes nonnegative regularization parameter, $\|\mathbf{y} - \mathbf{Ax}\|_F^2$ denotes approximation error, and $\|\mathbf{x}\|_F^2$ denotes solution size.

In [34], Oraintara et al. proposed an iterative algorithm to compute $\lambda$ based on the L-curve criterion depicted in fig. 1. In summary, $\mathbf{x}$ and $\lambda$ can be computed using algorithm 1:

---

**Algorithm 1** Iterative algorithm for computing $\mathbf{x}$ and $\lambda$.

---

Initialize $\mathbf{x}^{(0)}$ and $\lambda^{(0)}$.
Set $k \leftarrow 0$
**repeat**

$$k \leftarrow k + 1$$
$$\mathbf{x}^{(k)} \leftarrow \arg\min_{\mathbf{x}} \|\mathbf{y} - \mathbf{Ax}\|_F^2 + \lambda^{(k-1)} \|\mathbf{x}\|_F^2 \tag{7}$$
$$\lambda^{(k)} \leftarrow |\gamma| \frac{\|\mathbf{y} - \mathbf{Ax}^{(k)}\|_F^2}{\|\mathbf{x}^{(k)}\|_F^2} \tag{8}$$
$$\text{error} \leftarrow \frac{\|\mathbf{x}^{(k)} - \mathbf{x}^{(k-1)}\|}{\mathbf{x}^{(k-1)}}$$

**until** error $\leq \epsilon$

---

where $\gamma$ is the slope of the straight line that is tangent to the L-curve and $\epsilon$ is a small nonnegative number that is set to 0.001 in the authors' work [34]. Note that the value of $\gamma$ doesn't influence convergence property of sequence $\mathbf{x}^{(k)}$ and $\lambda^{(k)}$, and as long as $\lambda^{(0)}$ is sufficiently small then $\lambda^{(k)}$ converges to a stationary point [34]. The similar method for computing $\lambda$ can also be found in ref. [33], but the authors fixed $\gamma$ value to one.

We will now derive formulation for Tikhonov regularized NMF. The NMF problem in eq. 2 is known to be nonconvex and may have several local mimima [25]. The common practice to deal with the nonconvexity of an optimization problem is by transforming it into convex subproblems [41]. In the case of the NMF, this can be done by employing the alternating strategy; fixing one matrix while solving for the other [25] (apparently, all NMF algorithms utilizing alternating strategy). This strategy transforms an NMF problem into a pair of convex subproblems. The following equations give convex subproblems of the NMF,

$$\mathbf{B} = \arg\min_{\mathbf{B} \geq 0} \frac{1}{2}\|\mathbf{A} - \mathbf{BC}\|_F^2 \qquad (9)$$

$$\mathbf{C} = \arg\min_{\mathbf{C} \geq 0} \frac{1}{2}\|\mathbf{A} - \mathbf{BC}\|_F^2. \qquad (10)$$

Alternately solving eq. 9 for $\mathbf{B}$ and eq. 10 for $\mathbf{C}$ is known as alternating nonnegativity-constrained LS (ANLS), and usually each of these subproblems is solved by decomposing it into a series of corresponding nonnegativity-constraint LS (NNLS) problems. The following equations are the NNLS versions of eq. 9 and eq. 10.

$$\mathfrak{b}_m^T = \arg\min_{\mathfrak{b}_m \geq 0} \frac{1}{2}\|\mathfrak{a}_m^T - \mathbf{C}^T\mathfrak{b}_m^T\|_F^2, \quad \forall m \qquad (11)$$

$$\mathbf{c}_n = \arg\min_{\mathbf{c}_n \geq 0} \frac{1}{2}\|\mathbf{a}_n - \mathbf{Bc}_n\|_F^2, \quad \forall n, \qquad (12)$$

where $\mathfrak{b}_m$ and $\mathfrak{a}_m$ denotes the $m$-th row of $\mathbf{B}$ and $\mathbf{A}$ respectively.

As shown, each of these NNLS problems is exactly the standard LS problem with additional nonnegativity constraint. Accordingly, Tikhonov regularization can be employed to improve the solutions. The following equations give Tikhonov regularized version of the above NNLS problems.

$$\mathfrak{b}_m^T = \arg\min_{\mathfrak{b}_m \geq 0} \frac{1}{2}\|\mathfrak{a}_m^T - \mathbf{C}^T\mathfrak{b}_m^T\|_F^2 + \frac{1}{2}\beta_m\|\mathfrak{b}_m^T\|_F^2 \ \forall m, \quad (13)$$

$$\mathbf{c}_n = \arg\min_{\mathbf{c}_n \geq 0} \frac{1}{2}\|\mathbf{a}_n - \mathbf{Bc}_n\|_F^2 + \frac{1}{2}\alpha_n\|\mathbf{c}_n\|_F^2, \quad \forall n, \quad (14)$$

where $\alpha_n$ and $\beta_m$ denotes the corresponding nonnegative regularization parameters. By rearranging rows of $\mathbf{B}$ and columns of $\mathbf{C}$ back, *Tikhonov regularized NMF* can be written as:

$$\mathbf{B} = \arg\min_{\mathbf{B} \geq 0} \frac{1}{2}\|\mathbf{A} - \mathbf{BC}\|_F^2 + \frac{1}{2}\|\sqrt{\beta}\mathbf{B}\|_F^2, \quad (15)$$

$$\mathbf{C} = \arg\min_{\mathbf{C} \geq 0} \frac{1}{2}\|\mathbf{A} - \mathbf{BC}\|_F^2 + \frac{1}{2}\|\mathbf{C}\sqrt{\alpha}\|_F^2, \quad (16)$$

where $\boldsymbol{\beta} = \mathrm{diag}(\beta_1, \ldots, \beta_M)$ and $\boldsymbol{\alpha} = \mathrm{diag}(\alpha_1, \ldots, \alpha_N)$.

The following gives a generic algorithm for Tikhonov regularized NMF where update rules for $\alpha_n$ and $\beta_m$ are derived based on the work of Oraintara et al. [34], $\gamma_m^B$ and $\gamma_n^C$ are defined similarly as in algorithm 1, and $\epsilon$ denotes small positive number.

## III. A CONVERGED ALGORITHM FOR TIKHONOV REGULARIZED NMF

We will now present a converged algorithm for Tikhonov regularized NMF based on additive update rules. By combining update rules for $\mathbf{B}$ and $\mathbf{C}$ in eq. 15 and eq. 16, we define

---

**Algorithm 2** A generic algorithm for Tikhonov regularized NMF.

Initialize $\mathbf{B}^{(0)}$, $\mathbf{C}^{(0)}$, $\boldsymbol{\alpha}^{(0)}$, and $\boldsymbol{\beta}^{(0)}$.
Set $k \leftarrow 0$
**repeat**

$$\mathbf{B}^{(k+1)} \leftarrow \arg\min_{\mathbf{B} \geq 0} \frac{1}{2}\|\mathbf{A} - \mathbf{BC}^{(k)}\|_F^2 + \frac{1}{2}\|\sqrt{\boldsymbol{\beta}^{(k)}}\mathbf{B}\|_F^2$$

$$\mathbf{C}^{(k+1)} \leftarrow \arg\min_{\mathbf{C} \geq 0} \frac{1}{2}\|\mathbf{A} - \mathbf{B}^{(k+1)}\mathbf{C}\|_F^2 + \frac{1}{2}\|\mathbf{C}\sqrt{\boldsymbol{\alpha}^{(k)}}\|_F^2$$

$$\beta_m^{(k+1)} \leftarrow |\gamma_m^B| \frac{\|\mathfrak{a}_m^T - \mathbf{C}^{(k)T}\mathfrak{b}_m^{(k+1)T}\|_F^2}{\|\mathfrak{b}_m^{(k+1)T}\|_F^2}, \quad \forall m$$

$$\alpha_n^{(k+1)} \leftarrow |\gamma_n^C| \frac{\|\mathbf{a}_n - \mathbf{B}^{(k+1)}\mathbf{c}_n^{(k+1)}\|_F^2}{\|\mathbf{c}_n^{(k+1)}\|_F^2}, \quad \forall n$$

$$k \leftarrow k + 1$$

**until**

$$\max\left(\nabla_{\mathbf{B}}J(\mathbf{B}) \odot \mathbf{B}\right) \leq \epsilon \ \& \ \max\left(\nabla_{\mathbf{C}}J(\mathbf{C}) \odot \mathbf{C}\right) \leq \epsilon$$

---

objective function for Tikhonov regularized NMF as follows:

$$\min_{\mathbf{B},\mathbf{C}} J(\mathbf{B},\mathbf{C}) = \frac{1}{2}\|\mathbf{A} - \mathbf{BC}\|_F^2 + \frac{1}{2}\|\sqrt{\boldsymbol{\beta}}\mathbf{B}\|_F^2 + \frac{1}{2}\|\mathbf{C}\sqrt{\boldsymbol{\alpha}}\|_F^2 \quad (17)$$

s.t. $\mathbf{B} \geq 0, \mathbf{C} \geq 0$.

The Karush-Kuhn-Tucker (KKT) function of the objective can be written as:

$$L(\mathbf{B},\mathbf{C}) = J(\mathbf{B},\mathbf{C}) - \mathrm{tr}\left(\boldsymbol{\Gamma}_{\mathbf{B}}\mathbf{B}^T\right) - \mathrm{tr}\left(\boldsymbol{\Gamma}_{\mathbf{C}}\mathbf{C}\right).$$

where $\boldsymbol{\Gamma}_{\mathbf{B}} \in \mathbb{R}^{M \times R}$ and $\boldsymbol{\Gamma}_{\mathbf{C}} \in \mathbb{R}^{N \times R}$ denotes the KKT multipliers. Partial derivatives of $L$ with respect to $\mathbf{B}$ and $\mathbf{C}$ are:

$$\nabla_{\mathbf{B}}L(\mathbf{B}) = \nabla_{\mathbf{B}}J(\mathbf{B}) - \boldsymbol{\Gamma}_{\mathbf{B}},$$

$$\nabla_{\mathbf{C}}L(\mathbf{C}) = \nabla_{\mathbf{C}}J(\mathbf{C}) - \boldsymbol{\Gamma}_{\mathbf{C}}^T,$$

with

$$\nabla_{\mathbf{B}}J(\mathbf{B}) = \mathbf{BCC}^T - \mathbf{AC}^T + \boldsymbol{\beta}\mathbf{B},$$

$$\nabla_{\mathbf{C}}J(\mathbf{C}) = \mathbf{B}^T\mathbf{BC} - \mathbf{B}^T\mathbf{A} + \mathbf{C}\boldsymbol{\alpha}.$$

By results from optimization studies, $(\mathbf{B}^*, \mathbf{C}^*)$ is a stationary point of eq. 17 if it satisfies the KKT optimality conditions [42], i.e.,

$$\mathbf{B}^* \geq \mathbf{0}, \qquad\qquad \mathbf{C}^* \geq \mathbf{0}, \qquad (18)$$

$$\nabla_{\mathbf{B}}J(\mathbf{B}^*) = \boldsymbol{\Gamma}_{\mathbf{B}} \geq \mathbf{0}, \qquad \nabla_{\mathbf{C}}J(\mathbf{C}^*) = \boldsymbol{\Gamma}_{\mathbf{C}}^T \geq \mathbf{0}, \quad (19)$$

$$\nabla_{\mathbf{B}}J(\mathbf{B}^*) \odot \mathbf{B}^* = \mathbf{0}, \qquad \nabla_{\mathbf{C}}J(\mathbf{C}^*) \odot \mathbf{C}^* = \mathbf{0}, \quad (20)$$

where $\odot$ denotes Hadamard products (component-wise multiplications), and eq. 20 is known as the complementary slackness.

Multiplicative update rules based algorithm for Tikhonov regularized NMF can be derived by utilizing the complementary slackness:

$$(\mathbf{BCC}^T - \mathbf{AC}^T + \beta\mathbf{B}) \odot \mathbf{B} = \mathbf{0},$$
$$(\mathbf{B}^T\mathbf{BC} - \mathbf{B}^T\mathbf{A} + \mathbf{C}\alpha) \odot \mathbf{C} = \mathbf{0}.$$

These equations lead to the following update rules:

$$b_{mr} \longleftarrow b_{mr} \frac{(\mathbf{AC}^T)_{mr}}{(\mathbf{BCC}^T + \beta\mathbf{B})_{mr}} \quad \forall m, r \quad (21)$$

$$c_{rn} \longleftarrow c_{rn} \frac{(\mathbf{B}^T\mathbf{A})_{rn}}{(\mathbf{B}^T\mathbf{BC} + \mathbf{C}\alpha)_{rn}} \quad \forall r, n \quad (22)$$

where $b_{mr}$ and $c_{rn}$ denote $(m, r)$ entry of $\mathbf{B}$ and $(r, n)$ entry of $\mathbf{C}$ respectively.

As stated by Lin [18], the above multiplicative update rules based algorithm can be modified into an equivalent converged algorithm by (1) using additive update rules, and (2) replacing zero entries that do not satisfy the KKT conditions with a small positive number to escape the zero locking.

The additive update rules version of the algorithm can be written as:

$$b_{mr} \longleftarrow b_{mr} - \frac{b_{mr}}{(\mathbf{BCC}^T + \beta\mathbf{B})_{mr}} \nabla_{\mathbf{B}}J(\mathbf{B})_{mr},$$

$$c_{rn} \longleftarrow c_{rn} - \frac{c_{rn}}{(\mathbf{B}^T\mathbf{BC} + \mathbf{C}\alpha)_{rn}} \nabla_{\mathbf{C}}J(\mathbf{C})_{rn}.$$

By inspection it is clear that this algorithm also suffers from the zero locking, i.e.:

$$\nabla_{\mathbf{B}}J(\mathbf{B})_{mr} < 0 \ \& \ b_{mr} = 0, \quad \text{or}$$
$$\nabla_{\mathbf{C}}J(\mathbf{C})_{rn} < 0 \ \& \ c_{rn} = 0,$$

such that when these conditions happened, the algorithm can no longer update the corresponding entries even though those entries haven't satisfied the KKT optimality conditions in eq. 19.

Algorithm 3 gives necessary modifications to avoid zero locking and—as will be shown later—also has convergence guarantee, where $\oslash$ denotes component-wise division,

$$\bar{b}_{mr}^{(k)} \equiv \begin{cases} b_{mr}^{(k)} & \text{if } \nabla_{\mathbf{B}}J(\mathbf{B}^{(k)}, \mathbf{C}^{(k)})_{mr} \geq 0 \\ \max(b_{mr}^{(k)}, \sigma) & \text{if } \nabla_{\mathbf{B}}J(\mathbf{B}^{(k+1)}, \mathbf{C}^{(k)})_{mr} < 0 \end{cases}, \quad (23)$$

$$\bar{c}_{rn}^{(k)} \equiv \begin{cases} c_{rn}^{(k)} & \text{if } \nabla_{\mathbf{C}}J(\mathbf{B}^{(k+1)}, \mathbf{C}^{(k)})_{rn} \geq 0 \\ \max(c_{rn}^{(k)}, \sigma) & \text{if } \nabla_{\mathbf{C}}J(\mathbf{B}^{(k+1)}, \mathbf{C}^{(k)})_{rn} < 0 \end{cases}, \quad (24)$$

denote the modifications to avoid the zero locking with $\sigma$ is a small positive number, $\delta_{\mathbf{B}}$ and $\delta_{\mathbf{C}}$ denote small positive numbers that introduced to avoid division by zeros, $\bar{\mathbf{B}}$ and $\bar{\mathbf{C}}$ denote matrices that contain $\bar{b}_{mr}$ and $\bar{c}_{rn}$ respectively, and

$$\nabla_{\mathbf{B}}J(\mathbf{B}^{(k)}, \mathbf{C}^{(k)}) = \mathbf{B}^{(k)}\mathbf{C}^{(k)}\mathbf{C}^{(k)T} - \mathbf{AC}^{(k)T} + \beta^{(k)}\mathbf{B}^{(k)},$$

$$\nabla_{\mathbf{C}}J(\mathbf{B}^{(k+1)}, \mathbf{C}^{(k)}) = \mathbf{B}^{(k+1)T}\mathbf{B}^{(k+1)}\mathbf{C}^{(k)} - \mathbf{B}^{(k+1)T}\mathbf{A} + \mathbf{C}^{(k)}\alpha^{(k)}.$$

---

**Algorithm 3** A converged algorithm for Tikhonov regularized NMF.

Initialization $\mathbf{B}^{(0)} \geq \mathbf{0}$, $\mathbf{C}^{(0)} \geq \mathbf{0}$, $\beta_m^{(0)} \geq 0 \, \forall m$, and $\alpha_n^{(0)} \geq 0 \, \forall n$.
$k \leftarrow 0$
**repeat**

$$\mathbf{B}^{(k+1)} \leftarrow \mathbf{B}^{(k)} - \bar{\mathbf{B}}^{(k)} \odot \nabla_{\mathbf{B}}J(\mathbf{B}^{(k)}, \mathbf{C}^{(k)}) \oslash$$
$$(\bar{\mathbf{B}}^{(k)}\mathbf{C}^{(k)}\mathbf{C}^{(k)T} + \beta^{(k)}\bar{\mathbf{B}}^{(k)} + \delta_{\mathbf{B}}) \quad (25)$$

$$\mathbf{C}^{(k+1)} \leftarrow \mathbf{C}^{(k)} - \bar{\mathbf{C}}^{(k)} \odot \nabla_{\mathbf{C}}J(\mathbf{B}^{(k+1)}, \mathbf{C}^{(k)}) \oslash$$
$$(\mathbf{B}^{(k+1)T}\mathbf{B}^{(k+1)}\bar{\mathbf{C}}^{(k)} + \bar{\mathbf{C}}^{(k)}\alpha^{(k)} + \delta_{\mathbf{C}}) \quad (26)$$

$$\beta_m^{(k)} \leftarrow |\gamma_m^B| \frac{\|\mathfrak{a}_m^T - \mathbf{C}^{(k)T}\mathfrak{b}_m^{(k+1)T}\|_F^2}{\|\mathfrak{b}_m^{(k+1)T}\|_F^2 + \delta_{\mathbf{B}}}, \quad \forall m \quad (27)$$

$$\alpha_n^{(k+1)} \leftarrow |\gamma_n^C| \frac{\|\mathbf{a}_n - \mathbf{B}^{(k+1)}\mathbf{c}_n^{(k+1)}\|_F^2}{\|\mathbf{c}_n^{(k+1)}\|_F^2 + \delta_{\mathbf{C}}}, \quad \forall n \quad (28)$$

$k \leftarrow k + 1$
**until**

$$\max(\nabla_{\mathbf{B}}J(\mathbf{B}) \odot \mathbf{B}) \leq \epsilon \ \& \ \max(\nabla_{\mathbf{C}}J(\mathbf{C}) \odot \mathbf{C}) \leq \epsilon$$

---

where $0 < \text{step} < 1$. Note that since algorithm 3 is free from the zero locking, $\mathbf{B}$ and $\mathbf{C}$ can be initialized using nonnegative matrices.

*Theorem 1:* If $\mathbf{B}^0 > 0$ and $\mathbf{C}^0 > 0$, then $\mathbf{B}^k > 0$ and $\mathbf{C}^k > 0$, $\forall k \geq 0$. And if $\mathbf{B}^0 \geq 0$ and $\mathbf{C}^0 \geq 0$, then $\mathbf{B}^k \geq 0$ and $\mathbf{C}^k \geq 0$, $\forall k \geq 0$.

*Proof:* This statement is clear for $k = 0$, so we need only to prove for $k > 0$.

*Case 1:* $\nabla_{\mathbf{B}}J_{mr} \geq 0 \Rightarrow \bar{b}_{mr} = b_{mr}$ (see $\bar{b}_{mr}$ definition in eq. 23).

$$b_{mr}^{(k+1)} = \frac{(\mathbf{B}^{(k)}\mathbf{C}^{(k)}\mathbf{C}^{(k)T} + \beta^{(k)}\mathbf{B}^{(k)})_{mr}b_{mr}^{(k)} + \delta_{\mathbf{B}}b_{mr}^k}{(\mathbf{B}^{(k)}\mathbf{C}^{(k)}\mathbf{C}^{(k)T} + \beta^{(k)}\mathbf{B}^{(k)})_{mr} + \delta_{\mathbf{B}}} -$$
$$\frac{(\mathbf{B}^{(k)}\mathbf{C}^{(k)}\mathbf{C}^{(k)T} - \mathbf{AC}^{(k)T} + \beta^{(k)}\mathbf{B}^{(k)})_{mr}b_{mr}^{(k)}}{(\mathbf{B}^{(k)}\mathbf{C}^{(k)}\mathbf{C}^{(k)T} + \beta^{(k)}\mathbf{B}^{(k)})_{mr} + \delta_{\mathbf{B}}}$$
$$= \frac{(\delta_{\mathbf{B}} + \mathbf{AC}^{(k)T})_{mr}b_{mr}^{(k)}}{(\mathbf{B}^{(k)}\mathbf{C}^{(k)}\mathbf{C}^{(k)T} + \beta^{(k)}\mathbf{B}^{(k)})_{mr} + \delta_{\mathbf{B}}}.$$

Thus $\forall k > 0$, $b_{mr}^{(k)} > 0 \Rightarrow b_{mr}^{(k+1)} > 0 \, \forall m, r$, and $b_{mr}^{(k)} \geq 0 \Rightarrow b_{mr}^{(k+1)} \geq 0 \, \forall m, r$.

*Case 2:* $\nabla_{\mathbf{B}}J_{mr} < 0 \Rightarrow \bar{b}_{mr} \neq b_{mr}$.

$$b_{mr}^{(k+1)} = b_{mr}^{(k)} - \frac{\max(b_{mr}^{(k)}, \sigma)\nabla_{\mathbf{B}}J(\mathbf{B}^{(k)}, \mathbf{C}^{(k)})_{mr}}{(\bar{\mathbf{B}}^{(k)}\mathbf{C}^{(k)}\mathbf{C}^{(k)T} + \beta^{(k)}\bar{\mathbf{B}}^{(k)})_{mr} + \delta_{\mathbf{B}}}.$$

Because $\max(b_{mr}^{(k)}, \sigma) > 0$ and $\nabla_{\mathbf{B}}J(\mathbf{B}^{(k)}, \mathbf{C}^{(k)})_{mr} < 0$, then $\forall k > 0$, $b_{mr}^{(k)} > 0 \Rightarrow b_{mr}^{(k+1)} > 0 \, \forall m, r$, and $b_{mr}^{(k)} \geq 0 \Rightarrow b_{mr}^{(k+1)} > 0 \, \forall m, r$.

*Case 3*: $\nabla_{\mathbf{C}} J_{rn} \geq 0 \Rightarrow \bar{c}_{rn} = c_{rn}$.

$$
\begin{aligned}
c_{rn}^{(k+1)} =& \frac{\left(\mathbf{B}^{(k+1)T}\mathbf{B}^{(k+1)}\mathbf{C}^{(k)} + \mathbf{C}^{(k)}\boldsymbol{\alpha}^{(k)}\right)_{rn} c_{rn}^{(k)} + \delta_{\mathbf{C}} c_{rn}^{(k)}}{\left(\mathbf{B}^{(k+1)T}\mathbf{B}^{(k+1)}\mathbf{C}^{(k)} + \mathbf{C}^{(k)}\boldsymbol{\alpha}^{(k)}\right)_{rn} + \delta_{\mathbf{C}}} - \\
& \frac{\left(\mathbf{B}^{(k+1)T}\mathbf{B}^{(k+1)}\mathbf{C}^{(k)} - \mathbf{B}^{(k+1)T}\mathbf{A} + \mathbf{C}^{(k)}\boldsymbol{\alpha}^{(k)}\right)_{rn} c_{rn}^{(k)}}{\left(\mathbf{B}^{(k+1)T}\mathbf{B}^{(k+1)}\mathbf{C}^{(k)} + \mathbf{C}^{(k)}\boldsymbol{\alpha}^{(k)}\right)_{rn} + \delta_{\mathbf{C}}} \\
=& \frac{\left(\delta_{\mathbf{C}} + \mathbf{B}^{(k+1)T}\mathbf{A}\right)_{rn} c_{rn}^{(k)}}{\left(\mathbf{B}^{(k+1)T}\mathbf{B}^{(k+1)}\mathbf{C}^{(k)} + \mathbf{C}^{(k)}\boldsymbol{\alpha}^{(k)}\right)_{rn} + \delta_{\mathbf{C}}},
\end{aligned}
$$

Thus $\forall k > 0$, $c_{rn}^{k} > 0 \Rightarrow c_{rn}^{(k+1)} > 0 \; \forall r, n$, and $c_{rn}^{k} \geq 0 \Rightarrow c_{rn}^{(k+1)} \geq 0 \; \forall r, n$.

*Case 4*: $\nabla_{\mathbf{C}} J_{rn} < 0 \Rightarrow \bar{c}_{rn} \neq c_{rn}$.

$$
c_{rn}^{(k+1)} = c_{rn}^{(k)} - \frac{\max\left(c_{rn}^{(k)}, \sigma\right) \bar{\nabla}_{\mathbf{C}} J\left(\mathbf{B}^{(k+1)}, \mathbf{C}^{(k)}\right)_{rn}}{\left(\mathbf{B}^{(k+1)T}\mathbf{B}^{(k+1)}\bar{\mathbf{C}}^{(k)} + \bar{\mathbf{C}}^{(k)}\boldsymbol{\alpha}^{(k)}\right)_{rn} + \delta_{\mathbf{C}}}.
$$

Because $\max\left(c_{rn}^{(k)}, \sigma\right) > 0$ and $\nabla_{\mathbf{C}} J\left(\mathbf{B}^{(k+1)}, \mathbf{C}^{(k)}\right)_{rn} < 0$, then $\forall k > 0$, $c_{rn}^{(k)} > 0 \Rightarrow c_{rn}^{(k+1)} > 0 \; \forall r, n$, and $c_{rn}^{(k)} \geq 0 \Rightarrow c_{rn}^{(k+1)} > 0 \; \forall r, n$.

By combining results for $k = 0$ and $k > 0$ in case 1-4, the proof is completed. ∎

Appendix A gives Matlab/Octave codes for implementing algorithm 3.

## IV. Convergence Analysis

There are two type of update rules in algorithm 3. The first is the update rules for $\mathbf{B}^{(k)}$ and $\mathbf{C}^{(k)}$, and the second is the update rules for $\boldsymbol{\beta}^{(k)}$ and $\boldsymbol{\alpha}^{(k)}$. Since the algorithm 3 uses alternating strategy in updating these variables, the convergence analysis can be carried out separately. This approach is known as the *block-coordinate descent* method [42].

To derive the convergence guarantee of *solution sequence* $\{\mathbf{B}^{(k)}, \mathbf{C}^{(k)}, \boldsymbol{\beta}^{(k)}, \boldsymbol{\alpha}^{(k)}\}$, we will first show the convergence of sequence $\{\mathbf{B}^{(k)}, \mathbf{C}^{(k)}\}$, and then sequence $\{\boldsymbol{\beta}^{(k)}, \boldsymbol{\alpha}^{(k)}\}$. From convergence analysis study, the following conditions must be satisfied for sequence $\{\mathbf{B}^{(k)}, \mathbf{C}^{(k)}\}$ to have convergence guarantee [18], [25], [44].

1) The nonincreasing property of sequence $J\left(\mathbf{B}^{(k)}, \mathbf{C}^{(k)}\right)$, i.e.,
   a) $J\left(\mathbf{B}^{(k+1)}\right) \leq J\left(\mathbf{B}^{(k)}\right)$ and
   b) $J\left(\mathbf{C}^{(k+1)}\right) \leq J\left(\mathbf{C}^{(k)}\right)$.
2) Any limit point of sequence $\{\mathbf{B}^{(k)}, \mathbf{C}^{(k)}\}$ generated by algorithm 3 is a stationary point.
3) Sequence $\{\mathbf{B}^{(k)}, \mathbf{C}^{(k)}\}$ has at least one limit point.

### A. The nonincreasing property of sequence $J\left(\mathbf{B}^{(k)}\right)$

We will utilize the auxiliary function approach introduced in [1] to prove this property. By using the auxiliary function as an intermediate function, the nonincreasing property of $J\left(\mathbf{B}^{(k)}\right)$ can be restated with:

$$
\begin{aligned}
J\left(\mathbf{B}^{(k+1)}\right) &= G\left(\mathbf{B}^{(k+1)}, \mathbf{B}^{(k+1)}\right) \leq G\left(\mathbf{B}^{(k+1)}, \mathbf{B}^{(k)}\right) \\
&\leq G\left(\mathbf{B}^{(k)}, \mathbf{B}^{(k)}\right) = J\left(\mathbf{B}^{(k)}\right).
\end{aligned}
$$

To define $G$, let's rearrange $\mathbf{B}$ into:

$$
\mathfrak{B}^T \equiv \begin{bmatrix} \mathfrak{b}_1^T & & & \\ & \mathfrak{b}_2^T & & \\ & & \ddots & \\ & & & \mathfrak{b}_M^T \end{bmatrix}
$$

where $\mathfrak{b}_m$ denotes the $m$-th row of $\mathbf{B}$. And also let's define:

$$
\nabla_{\mathfrak{B}^T}\mathfrak{J}\left(\mathfrak{B}^{kT}\right) \equiv \begin{bmatrix} \nabla_{\mathbf{B}}\mathfrak{J}\left(\mathbf{B}^{(k)}\right)_1^T & & \\ & \ddots & \\ & & \nabla_{\mathbf{B}}\mathfrak{J}\left(\mathbf{B}^{(k)}\right)_M^T \end{bmatrix}
$$

where $\nabla_{\mathbf{B}}\mathfrak{J}\left(\mathbf{B}^{(k)}\right)_m$ denotes the $m$-th row of $\nabla_{\mathbf{B}} J\left(\mathbf{B}^{(k)}\right) = \mathbf{B}^{(k)}\mathbf{C}^{(k)}\mathbf{C}^{(k)T} - \mathbf{A}\mathbf{C}^{(k)T} + \boldsymbol{\beta}^{(k)}\mathbf{B}^{(k)}$. Then define,

$$
\mathbf{D} \equiv \text{diag}\left(\mathbf{D}^1, \dots, \mathbf{D}^M\right)
$$

where $\mathbf{D}^m$ denotes a diagonal matrix with its diagonal entries defined as:

$$
d_{rr}^m \equiv \begin{cases} \frac{\left(\bar{\mathbf{B}}^{(k)}\mathbf{C}^{(k)}\mathbf{C}^{(k)T} + \boldsymbol{\beta}^{(k)}\bar{\mathbf{B}}^{(k)}\right)_{mr} + \delta_{\mathbf{B}}}{\bar{b}_{mr}^{(k)}} & \text{if } r \in \mathcal{I}_m \\ \star & \text{if } r \notin \mathcal{I}_m \end{cases}
$$

with

$$
\begin{aligned}
\mathcal{I}_m \equiv \big\{ r | b_{mr}^{(k)} > 0,\ \nabla_{\mathbf{B}} J\left(\mathbf{B}^{(k)}\right)_{mr} \neq 0,\ \text{or} \\
b_{mr}^{(k)} = 0,\ \nabla_{\mathbf{B}} J\left(\mathbf{B}^{(k)}\right)_{mr} < 0 \big\}
\end{aligned}
$$

denotes the set of non-KKT indices in $m$-th row of $\mathbf{B}^{(k)}$, and $\star$ is defined so that $\star \equiv 0$ and $\star^{-1} \equiv 0$.

The auxiliary function $\mathfrak{G}$ can be defined as:

$$
\begin{aligned}
\mathfrak{G}\left(\mathfrak{B}^T, \mathfrak{B}^{(k)T}\right) \equiv& \mathfrak{J}\left(\mathfrak{B}^{(k)T}\right) + \text{tr}\left\{\left(\mathfrak{B} - \mathfrak{B}^{(k)}\right)\nabla_{\mathfrak{B}^T}\mathfrak{J}\left(\mathfrak{B}^{(k)T}\right)\right\} \\
&+ \frac{1}{2}\text{tr}\left\{\left(\mathfrak{B} - \mathfrak{B}^{(k)}\right)\mathbf{D}\left(\mathfrak{B} - \mathfrak{B}^{(k)}\right)^T\right\}.
\end{aligned} \tag{29}
$$

Note that $\mathfrak{J}$ and $\mathfrak{G}$ are equivalent to $J$ and $G$ with $\mathbf{B}$ is rearranged into $\mathfrak{B}^T$, and other variables are reordered accordingly. And also whenever $\mathbf{X}^{(k+1)}$ is a variable, we remove $(k+1)$ sign. And:

$$
\nabla_{\mathfrak{B}^T}\mathfrak{G}\left(\mathfrak{B}^T, \mathfrak{B}^{(k)T}\right) = \mathbf{D}\left(\mathfrak{B} - \mathfrak{B}^{(k)}\right)^T + \nabla_{\mathfrak{B}^T}\mathfrak{J}\left(\mathfrak{B}^{(k)T}\right).
$$

By definition $\mathbf{D}$ is positive definite for all $\mathbf{B}^{(k)}$ not satisfy the KKT conditions and positive semidefinite if and only if $\mathbf{B}^{(k)}$ satisfies the KKT conditions. Thus $\mathfrak{G}\left(\mathfrak{B}^T, \mathfrak{B}^{(k)T}\right)$ is a strict convex function, and consequently has a unique minimum, so that:

$$
\begin{aligned}
\mathbf{D}\left(\mathfrak{B} - \mathfrak{B}^{(k)}\right)^T + \nabla_{\mathfrak{B}^T}\mathfrak{J}\left(\mathfrak{B}^{(k)T}\right) = 0, \\
\mathfrak{B}^T = \mathfrak{B}^{(k)T} - \mathbf{D}^{-1}\nabla_{\mathfrak{B}^T}\mathfrak{J}\left(\mathfrak{B}^{(k)T}\right),
\end{aligned} \tag{30}
$$

which is exactly the update rule for $\mathbf{B}$ in eq. 25.

*Lemma 1*: $\mathfrak{J}\left(\mathfrak{B}^T\right)$ can be rewritten as:

$$
\begin{aligned}
\mathfrak{J}\left(\mathfrak{B}^T\right) =& \mathfrak{J}\left(\mathfrak{B}^{(k)T}\right) + \text{tr}\left\{\left(\mathfrak{B} - \mathfrak{B}^{(k)}\right)\nabla_{\mathfrak{B}^T}\mathfrak{J}\left(\mathfrak{B}^{(k)T}\right)\right\} \\
&+ \frac{1}{2}\text{tr}\left\{\left(\mathfrak{B} - \mathfrak{B}^{(k)}\right)\nabla_{\mathbf{B}}^2\mathbf{J}\left(\mathbf{B}^{(k)}\right)\left(\mathfrak{B} - \mathfrak{B}^{(k)}\right)^T\right\}.
\end{aligned} \tag{31}
$$

where

$$\nabla_{\mathbf{B}}^2 \mathbf{J}(\mathbf{B}^k) \equiv \begin{bmatrix} \mathbf{C}^{(k)}\mathbf{C}^{(k)T} + \beta_1 \mathbf{I} & & \\ & \ddots & \\ & & \mathbf{C}^{(k)}\mathbf{C}^{(k)T} + \beta_M \mathbf{I} \end{bmatrix}$$

and $\mathbf{I}$ denotes corresponding compatible identity matrix.

*Proof:* Let decompose $J(\mathbf{B})$ into:

$$J(\mathbf{B})_m = \frac{1}{2}\|\mathfrak{a}_m^T - \mathbf{C}^{(k)T}\mathfrak{b}_m^T\|_F^2 + \frac{1}{2}\beta_m\|\mathfrak{b}_m^T\|_F^2 \ \forall m,$$

so that $J(\mathbf{B}) = J(\mathbf{B})_1 \dots J(\mathbf{B})_M$. Then,

$$\frac{\partial J(\mathbf{B})_m}{\partial \mathfrak{b}_m} = -\mathfrak{a}_m\mathbf{C}^{(k)T} + \mathfrak{b}_m\mathbf{C}^{(k)}\mathbf{C}^{(k)T} + \beta_m\mathfrak{b}_m$$

and

$$\frac{\partial^2 J(\mathbf{B})_m}{\partial \mathfrak{b}_m^2} = \mathbf{C}^{(k)}\mathbf{C}^{(k)T} + \beta_m\mathbf{I}.$$

By using the Taylor series expansion, $J(\mathbf{B})_m$ can be rewritten as:

$$J(\mathbf{B})_m = J(\mathbf{B}^{(k)})_m + (\mathfrak{b}_m - \mathfrak{b}_m^{(k)}) \left(\frac{\partial J(\mathbf{B})_m}{\partial \mathfrak{b}_m}\right)^T$$
$$+ \frac{1}{2}(\mathfrak{b}_m - \mathfrak{b}_m^{(k)}) \left(\frac{\partial^2 J(\mathbf{B})_m}{\partial \mathfrak{b}_m^2}\right)(\mathfrak{b}_m - \mathfrak{b}_m^{(k)})^T,$$

which is the $m$-th row of $\mathfrak{J}(\mathfrak{B}^T)$. ∎

To prove the nonincreasing property of $J(\mathbf{B}^{(k)})$, the following statements must be shown:

1) $\mathfrak{G}(\mathfrak{B}^T, \mathfrak{B}^T) = \mathfrak{J}(\mathfrak{B}^T)$,
2) $\mathfrak{G}(\mathfrak{B}^{kT}, \mathfrak{B}^{kT}) = \mathfrak{J}(\mathfrak{B}^{kT})$,
3) $\mathfrak{G}(\mathfrak{B}^T, \mathfrak{B}^T) \leq \mathfrak{G}(\mathfrak{B}^T, \mathfrak{B}^{kT})$, and
4) $\mathfrak{G}(\mathfrak{B}^T, \mathfrak{B}^{kT}) \leq \mathfrak{G}(\mathfrak{B}^{kT}, \mathfrak{B}^{kT})$.

The first and second will be proven in theorem 2, the third in theorem 3, and the fourth in theorem 4.

*Theorem 2:* $\mathfrak{G}(\mathfrak{B}^T, \mathfrak{B}^T) = \mathfrak{J}(\mathfrak{B}^T)$ and $\mathfrak{G}(\mathfrak{B}^{(k)T}, \mathfrak{B}^{(k)T}) = \mathfrak{J}(\mathfrak{B}^{(k)T})$.

*Proof:* These are obvious from the definition of $\mathfrak{G}$ in eq. 29. ∎

*Theorem 3:* $\mathfrak{G}(\mathfrak{B}^T, \mathfrak{B}^T) \leq \mathfrak{G}(\mathfrak{B}^T, \mathfrak{B}^{(k)T})$. Moreover if and only if $\mathbf{B}^k$ satisfies the KKT conditions in eq. 18–20, then $\mathfrak{G}(\mathfrak{B}^T, \mathfrak{B}^T) = \mathfrak{G}(\mathfrak{B}^T, \mathfrak{B}^{(k)T})$.

*Proof:*

$$\mathfrak{G}(\mathfrak{B}^T, \mathfrak{B}^{(k)T}) - \mathfrak{G}(\mathfrak{B}^T, \mathfrak{B}^T) =$$
$$\frac{1}{2}\text{tr}\left\{(\mathfrak{B} - \mathfrak{B}^{(k)})\left(\mathbf{D} - \nabla_{\mathbf{B}}^2\mathbf{J}(\mathbf{B}^{(k)})\right)(\mathfrak{B} - \mathfrak{B}^{(k)})^T\right\} =$$
$$\frac{1}{2}\sum_{m=1}^{M}\left[(\mathfrak{b}_m - \mathfrak{b}_m^{(k)})\left(\mathbf{D}^m - \frac{\partial^2 J(\mathbf{B})_m}{\partial \mathfrak{b}_m^2}\right)(\mathfrak{b}_m - \mathfrak{b}_m^{(k)})^T\right]$$

If $\mathbf{D}^m - \frac{\partial^2 J(\mathbf{B})_m}{\partial \mathfrak{b}_m^2} \ \forall m$ are all positive definite, then the inequality always holds except when $\mathfrak{b}_m = \mathfrak{b}_m^k \ \forall m$, where based on theorem 1 and update rule eq. 25 happened if and only if the point has reach a stationary point—a point where the KKT conditions are satisfied. Thus, it is sufficient to prove the positive definiteness of $\mathbf{D}^m - \nabla_{\mathbf{B}}^2 J(\mathbf{B}^k)_m \ \forall m$.

Let $\mathbf{v}_m^T = \mathfrak{b}_m - \mathfrak{b}_m^{(k)} \neq \mathbf{0}$, then we must prove:

$$\mathbf{v}_m^T\left(\mathbf{D}^m - \frac{\partial^2 J(\mathbf{B})_m}{\partial \mathfrak{b}_m^2}\right)\mathbf{v}_m > 0.$$

Note that

$$d_{rr}^m \equiv \begin{cases} \frac{(\mathfrak{b}_m^{(k)}\mathbf{X}_m^{(k)})_r + \delta_{\mathbf{B}}}{\bar{b}_{mr}^{(k)}} & \text{if } r \in \mathcal{I}_m \\ \star & \text{if } r \notin \mathcal{I}_m \end{cases}$$

where $\bar{b}_m^{(k)}$ denotes the $m$-th row of $\bar{\mathbf{B}}^{(k)}$; and $\mathbf{X}_m^{(k)} = \mathbf{C}^{(k)}\mathbf{C}^{(k)T} + \beta_m\mathbf{I}$ and $\mathbf{D}^m$ are both symmetric matrix. Thus,

$$\mathbf{v}_m^T\left(\mathbf{D}^m - \frac{\partial^2 J(\mathbf{B})_m}{\partial \mathfrak{b}_m^2}\right)\mathbf{v}_m =$$
$$\sum_{r=1}^{R} v_r^2 \frac{\delta_{\mathbf{B}}}{\bar{b}_{mr}^{(k)}} + \sum_{r=1}^{R} v_r^2 \frac{(\mathbf{X}^{(k)}\bar{\mathbf{b}}_m^{(k)T})_r}{\bar{b}_{mr}^{(k)}} - \sum_{r,s=1}^{R} v_r v_s x_{rs}^{(k)} >$$
$$\sum_{r=1}^{R} v_r^2 \frac{\sum_{s=1}^{R} x_{rs}^{(k)}(\bar{\mathbf{b}}_m^{(k)})_s}{\bar{b}_{mr}^{(k)}} - \sum_{r=1}^{R}\sum_{s=1}^{R} v_r v_s x_{rs}^{(k)} =$$
$$\frac{1}{2}\sum_{r=1}^{R}\sum_{s=1}^{R} v_r^2 \frac{x_{rs}^{(k)}(\bar{\mathbf{b}}_m^{(k)})_s}{\bar{b}_{mr}^{(k)}} + \frac{1}{2}\sum_{s=1}^{R}\sum_{r=1}^{R} v_s^2 \frac{x_{sr}^{(k)}(\bar{\mathbf{b}}_m^{(k)})_r}{\bar{b}_{ms}^{(k)}}$$
$$- \sum_{r=1}^{R}\sum_{s=1}^{R} v_r v_s x_{rs}^{(k)} =$$
$$\frac{1}{2}\sum_{r=1}^{R}\sum_{s=1}^{R} x_{rs}^{(k)} \left(\sqrt{\frac{\bar{b}_{ms}^{(k)}}{\bar{b}_{mr}^{(k)}}} v_r - \sqrt{\frac{\bar{b}_{mr}^{(k)}}{\bar{b}_{ms}^{(k)}}} v_s\right)^2 \geq 0$$

where $v_r$ denotes the $r$-th entry of $\mathbf{v}_m$ and $x_{rs}^{(k)}$ denotes the $(r, s)$ entry of $\mathbf{X}^{(k)}$. Therefore, $\mathbf{D}^m - \frac{\partial^2 J(\mathbf{B})_m}{\partial \mathfrak{b}_m^2} \ \forall m$ are all positive definite ∎

*Theorem 4:* $\mathfrak{G}(\mathfrak{B}^T, \mathfrak{B}^{(k)T}) \leq \mathfrak{G}(\mathfrak{B}^{(k)T}, \mathfrak{B}^{(k)T})$. Moreover, if and only if $\mathbf{B}$ satisfies the KKT conditions, then $\mathfrak{G}(\mathfrak{B}^T, \mathfrak{B}^{(k)T}) = \mathfrak{G}(\mathfrak{B}^{(k)T}, \mathfrak{B}^{(k)T})$.

*Proof:*

$$\mathfrak{G}\left(\mathfrak{B}^{(k)T}, \mathfrak{B}^{(k)T}\right) - \mathfrak{G}\left(\mathfrak{B}^T, \mathfrak{B}^{(k)T}\right) =$$
$$- \text{tr}\left\{(\mathfrak{B} - \mathfrak{B}^{(k)})\nabla_{\mathfrak{B}^T}\mathfrak{J}(\mathfrak{B}^{(k)T})\right\}$$
$$- \frac{1}{2}\text{tr}\left\{(\mathfrak{B} - \mathfrak{B}^{(k)})\mathbf{D}(\mathfrak{B} - \mathfrak{B}^{(k)})^T\right\}.$$

By using eq. 30, and the fact that $\mathbf{D}$ is positive semi-definite:

$$\mathfrak{G}\left(\mathfrak{B}^{(k)T}, \mathfrak{B}^{(k)T}\right) - \mathfrak{G}\left(\mathfrak{B}^T, \mathfrak{B}^{(k)T}\right) =$$
$$\frac{1}{2}\text{tr}\left\{(\mathfrak{B} - \mathfrak{B}^{(k)})\mathbf{D}(\mathfrak{B} - \mathfrak{B}^{(k)})^T\right\} \geq 0,$$

we proved that $\mathfrak{G}(\mathfrak{B}^T, \mathfrak{B}^{(k)T}) \leq \mathfrak{G}(\mathfrak{B}^{(k)T}, \mathfrak{B}^{(k)T})$.

Now, let's prove the second part of the theorem. By the update rule eq. 25, if $\mathbf{B}^{(k)}$ satisfies the KKT conditions, then $\mathbf{B}$ will be equal to $\mathbf{B}^{(k)}$, and thus the equality holds. Now we need to prove that if the equality holds, then $\mathbf{B}^{(k)}$ satisfies the KKT conditions. To prove this, let consider a contradiction where the equality holds but $\mathbf{B}^{(k)}$ does not satisfy the KKT conditions. In this case, there exists at least an index $(m, r)$ such that:

$$b_{mr} \neq b_{mr}^{(k)} \text{ and } d_{rr}^m = \frac{\left(\bar{\mathbf{b}}_m^{(k)} \mathbf{X}_m^{(k)}\right)_r + \delta_{\mathbf{B}}}{\bar{b}_{mr}^{(k)}} \geq \frac{\delta_{\mathbf{B}}}{\bar{b}_{mr}^{(k)}}.$$

Note that by the definition in eq. 23, if $\bar{b}_{mr}^{(k)} = 0$, then it satisfies the KKT conditions. Accordingly, $b_{mr} = b_{mr}^{(k)}$ which violates the condition for contradiction. So, $\bar{b}_{mr}^k$ cannot be equal to zero, and thus $d_{rr}^m$ is well defined. Consequently,

$$\mathfrak{G}\left(\mathfrak{B}^{(k)T}, \mathfrak{B}^{(k)T}\right) - \mathfrak{G}\left(\mathfrak{B}^T, \mathfrak{B}^{(k)T}\right) \geq \frac{\left(b_{mr} - b_{mr}^{(k)}\right)^2 \delta_{\mathbf{B}}}{\bar{b}_{mr}^{(k)}} > 0,$$

which violates the condition for contradiction. Thus, it is proven that if the equality holds, then $\mathbf{B}^{(k)}$ satisfies the KKT conditions. ∎

The following theorem summarizes the nonincreasing property of sequence $J\left(\mathbf{B}^{(k)}\right)$

*Theorem 5:* $J\left(\mathbf{B}^{(k+1)}\right) \leq J\left(\mathbf{B}^{(k)}\right)$ $\forall k \geq 0$ under update rule eq. 25 with the equality happens if and only if $\mathbf{B}^{(k)}$ satisfies the KKT optimality conditions in eq. 18-20.

*Proof:* This is the results of theorem 2, 3, and 4. ∎

### B. The nonincreasing property of sequence $J\left(\mathbf{C}^{(k)}\right)$

*Theorem 6:* $J\left(\mathbf{C}^{(k+1)}\right) \leq J\left(\mathbf{C}^{(k)}\right)$ $\forall k \geq 0$ under update rule eq. 26 with the equality happens if and only if $\mathbf{C}^{(k)}$ satisfies the KKT optimality conditions in eq. 18-20.

*Proof:* This theorem can be proven similarly as in $J\left(\mathbf{B}^{(k)}\right)$ case. ∎

### C. The nonincreasing property of sequence $J\left(\mathbf{B}^{(k)}, \mathbf{C}^{(k)}\right)$

*Theorem 7:* $J\left(\mathbf{B}^{(k+1)}, \mathbf{C}^{(k+1)}\right) \leq J\left(\mathbf{B}^{(k+1)}, \mathbf{C}^{(k)}\right) \leq J\left(\mathbf{B}^{(k)}, \mathbf{C}^{(k)}\right)$ $\forall k \geq 0$ under update rule eq. 25 and 26 with the equality happens if and only if $\left(\mathbf{B}^{(k)}, \mathbf{C}^{(k)}\right)$ satisfies the KKT optimality conditions in eq. 18-20.

*Proof:* This statement can be proven by combining the results in theorem 5 and 6. ∎

By this theorem, we have proven the first conditions for the algorithm 3 to have convergence guarantee. The next subsection will give proofs for the second and the third condition.

### D. Limit points of sequence $\left\{\mathbf{B}^{(k)}, \mathbf{C}^{(k)}\right\}$

*Theorem 8:* Any limit point of sequence $\left\{\mathbf{B}^{(k)}, \mathbf{C}^{(k)}\right\}$ generated by algorithm 3 is a stationary point

*Proof:* By theorem 7, algorithm 3 produces strictly decreasing sequence $J\left(\mathbf{B}^{(k)}, \mathbf{C}^{(k)}\right)$ until reaching a point that satisfies the KKT conditions. By update rules in eq. 25 and 26, after a point satisfies the KKT conditions, the algorithm will stop updating $\left(\mathbf{B}^{(k)}, \mathbf{C}^{(k)}\right)$, i.e., $\mathbf{B}^{(k+1)} = \mathbf{B}^{(k)}$ and $\mathbf{C}^{(k+1)} = \mathbf{C}^{(k)}$ $\forall k \geq *$, where $*$ is the iteration where the KKT conditions are satisfied. ∎

*Theorem 9:* Sequence $\left\{\mathbf{B}^{(k)}, \mathbf{C}^{(k)}\right\}$ has at least one limit point.

*Proof:* As stated by Lin [18], it suffices to prove that $\left\{\mathbf{B}^{(k)}, \mathbf{C}^{(k)}\right\}$ is in a closed and bounded set. The boundedness of this sequence is clear by the objective eq. 17; if there exists $l$ such that $\lim b_{mr}^l \to \infty$ or $\lim c_{rn}^l \to \infty$, then $\lim J(\mathbf{B}^{(l)}, \mathbf{C}^{(l)}) \to \infty > J(\mathbf{B}^{(0)}, \mathbf{C}^{(0)})$ which violates theorem 7. With nonnegativity guarantee from theorem 1, it is proven that $\left\{\mathbf{B}^{(k)}, \mathbf{C}^{(k)}\right\}$ is in closed and bounded set. ∎

### E. The convergence of sequence $\left\{\boldsymbol{\beta}^{(k)}, \boldsymbol{\alpha}^{(k)}\right\}$

The convergence guarantee of this sequence has been established by Oraintara et al. in ref. [34], [43]. Here we will adopt their works and summarize the results in accord to our notations.

*Theorem 10 (Oraintara et al. [34]):* The optimum $\beta_m$ corresponding to the L-corner must satisfy

$$\beta_m \|\mathfrak{b}_m^T\|_F^2 = \left|\gamma_m^B\right| \|\mathfrak{a}_m^T - \mathbf{C}^T \mathfrak{b}_m^T\|_F^2.$$

The similar condition can also derived for $\alpha_n$.

Since the update rules for $\beta_m^{(k)}$ $\forall m$ and $\alpha_n^{(k)}$ $\forall n$ are derived from this theorem, it implies that the update rules find the optimal solutions of these parameters for each iteration.

Next we state the monotonicity property of sequence $\beta_m^{(k)}$ and $\alpha_m^{(k)}$.

*Lemma 2 (Oraintara et al. [34]):* The values of $\beta_m^{(k)}$ $\forall m$ either strictly increase or decrease under update rule eq. 27 unless converged to a limit point. The similar conditions also apply to $\alpha_n^{(k)}$ $\forall n$.

And, the following theorem summarizes the convergence property of sequence $\beta_m^{(k)}$ and $\alpha_n^{(k)}$.

*Theorem 11 (Oraintara et al. [34]):* If the update rule eq. 27 converges, then it converges to the corresponding L-corner. The same condition also applies to the update rule eq. 28

Thus, lemma 2 and theorem 11 state that while update rules eq. 27 and 28 generate monotonic updated values for $\beta_m^{(k)}$ $\forall m$ and $\alpha_n^{(k)}$ $\forall n$ which if the sequences converge, then they converge to the corresponding L-corner; there is no guarantee that the sequences will converge. Fortunately, the convergence can be established by choosing appropriate initial values.

Directly from ref. [34], the followings summarize the strategy in choosing the initial values.

1) If $\mathbf{a}_n$ is not in the range $\mathbf{B}$, then choosing $\alpha_n^{(0)} = 0$ will produce an increasing sequence of $\alpha_n^{(k)}$ converging to the nearest L-corner.

2) If $\mathbf{a}_n$ is in the range of $\mathbf{B}$, then it is always possible to choose $\alpha_n^{(0)} > 0$ small enough so that $J(\alpha_n^{(0)})_n > \alpha_n^{(0)}$ and convergence occurs.

3) More generally, if $\lim_{\alpha_n \to \infty} J(\alpha_n)_n/\alpha_n < 1$ and $J(0) > 0$, any initial value $\alpha_n^{(0)}$ produces a converging sequence $\alpha_n^{(k)}$.

4) When $\lim_{\alpha_n \to \infty} J(\alpha_n)_n/\alpha_n > 1$, the only case when the update rule eq. 28 will not generate a converged sequence is when $\alpha_n^{(0)}$ is chosen larger than the last intersection between $J(\alpha_n)_n$ and the straight line $\alpha_n = J(\alpha_n)_n$.

The same conditions can also be derived for sequence $\beta_m^{(k)}$ $\forall m$. Thus, by simply choosing $\alpha_n = 0$ $\forall n$ and $\beta_m = 0$ $\forall m$, the update rules eq. 27 and 28 will generate sequence $\beta_m^{(k)}$ $\forall m$ and $\alpha_n^{(k)}$ $\forall n$ that converge to the corresponding L-corners.

### F. The convergence of the solution sequence

The following theorem gives the convergence guarantee of the solution sequence $\{\mathbf{B}^{(k)}, \mathbf{C}^{(k)}, \boldsymbol{\beta}^{(k)}, \boldsymbol{\alpha}^{(k)}\}$ generated by algorithm 3

*Theorem 12:* By choosing appropriate initial values $\beta_m^{(0)}$ $\forall m$ and $\alpha_n^{(0)}$ $\forall n$, algorithm 3 generates sequence $\{\mathbf{B}^{(k)}, \mathbf{C}^{(k)}\}$ that converges to a point that satisfies the KKT optimality conditions (a stationary point), and sequence $\{\boldsymbol{\beta}^{(k)}, \boldsymbol{\alpha}^{(k)}\}$ that converges to the corresponding L-corners.

*Proof:* By the results of theorem 7, 8, and 9, we know that sequence $\{\mathbf{B}^{(k)}, \mathbf{C}^{(k)}\}$ converges to a point that satisfies the KKT conditions. By discussion in subsection IV-E we know that it is always possible to choose appropriate initial values for $\beta_m^{(0)}$ $\forall m$ and $\alpha_n^{(0)}$ $\forall n$ so that sequence $\{\boldsymbol{\beta}^{(k)}, \boldsymbol{\alpha}^{(k)}\}$ converges to the corresponding L-corners. ∎

## V. Discussion on the convergence of algorithm 3

Algorithm 3 converges when it stops updating both sequence $\{\mathbf{B}^{(k)}, \mathbf{C}^{(k)}\}$ and sequence $\{\boldsymbol{\beta}^{(k)}, \boldsymbol{\alpha}^{(k)}\}$. And as shown in eq. 27 and eq. 28, when $\{\mathbf{B}^{(k)}, \mathbf{C}^{(k)}\}$ has converged then $\{\boldsymbol{\beta}^{(k)}, \boldsymbol{\alpha}^{(k)}\}$ would also have converged. With the boundedness of $\mathbf{B}^{(k)}$ and $\mathbf{C}^{(k)}$, $\boldsymbol{\beta}^{(k)}$ and $\boldsymbol{\alpha}^{(k)}$ will also be bounded. Thus, algorithm 3 will be well-behaved through the update steps. This is an important fact since even though sequence $J(\mathbf{B}^{(k)}, \mathbf{C}^{(k)})$ has the nonincreasing property, due to lemma 2, algorithm 3 may produce nondecreasing $J(\mathbf{B}^{(k)}, \mathbf{C}^{(k)}, \boldsymbol{\beta}^{(k)}, \boldsymbol{\alpha}^{(k)})$ (we will refer this sequence as $J(\cdot)$). This is because the nonincreasing property of $J(\cdot)$ should be shown by proofing that:

1) $J(\mathbf{B}^{(k+1)}) \leq J(\mathbf{B}^{(k)})$,
2) $J(\mathbf{C}^{(k+1)}) \leq J(\mathbf{C}^{(k)})$,
3) $J(\boldsymbol{\beta}^{(k+1)}) \leq J(\boldsymbol{\beta}^{(k)})$, and
4) $J(\boldsymbol{\alpha}^{(k+1)}) \leq J(\boldsymbol{\alpha}^{(k)})$.

The first and second have been proven in theorem 5 and 6 respectively. But as stated in lemma 2, sequence $\beta_m$ $\forall m$ and $\alpha_n$ $\forall n$ can either strictly increase or decrease until reaching the limit points, thus $J(\boldsymbol{\beta}^{(k+1)}) > J(\boldsymbol{\beta}^{(k)})$ and/or $J(\boldsymbol{\alpha}^{(k+1)}) > J(\boldsymbol{\alpha}^{(k)})$ cases can occur. This, however, will not affect the convergence of algorithm 3 since as long as sequence $\{\boldsymbol{\beta}^{(k)}, \boldsymbol{\alpha}^{(k)}\}$ converges, then the update rules eq. 25 and 26 will eventually find the stationary point for $\{\mathbf{B}^{(k)}, \mathbf{C}^{(k)}\}$.

Numerically, it seems that $\gamma_m^B$ and $\gamma_n^C$ play the key role in determining whether sequence $\{\beta_m^{(k)}\}$ and $\{\alpha_n^{(k)}\}$ will strictly increase or decrease with the big values lead to the increasing sequences and vice versa.

## VI. Conclusion

We have presented a converged algorithm for NMF with Tikhonov regularization on its factor. There are two contributions that can be pointed out. The first is to show the connection between Tikhonov regularized linear inverse problems with constraint NMF which naturally leads to a mechanism for determining the regularization parameter in the NMF automatically. And, the second is the development of a converged algorithm for NMF with Tikhonov regularized constraints.

## Appendix A
## Octave/Matlab codes for algorithm 3

```
function [B, C, newalpha, newbeta, iteration,
    olderror, errordiff, maxNablaB,
    maxNablaC, resNorm, solNorm] =
    TikhonovNMF3(A, r, B0, C0, oldalpha,
    oldbeta, gammaB, gammaC, maxiter, tol)

%The converged version of the algorithm
%Use complementary slackness as stopping
    criterion

format long;
%%Check the input matrix
if min(min(A)) < 0
  error('Input matrix cannot contain negative
      entries');
  return
end

[m,n] = size(A);

%%Check input arguments

if ~exist('A')
  error('incorrect inputs!')
end
if ~exist('r')
  error('incorrect inputs!')
end
if ~exist('B0')
  B0 = rand(m,r);
end
if ~exist('C0')
  C0 = rand(r,n);
end
if ~exist('alpha0')
  oldalpha = zeros(n,1);
end
if ~exist('beta0')
  oldbeta = zeros(m,1);
end
if ~exist('gammaB')
  gammaB = ones(m,1)*0.1; %small values lead
      to better convergence property
end
if ~exist('gammaC')
  gammaC = ones(n,1)*0.1; %small values lead
      to better convergence property
end
if ~exist('maxiter')
  maxiter = 1000;
end
if ~exist('tol')
  tol = 1.0e-9;
end

B = B0; clear B0;
C = C0; clear C0;
newalpha = oldalpha;
newbeta = oldbeta;
trAtA = trace(A'*A);
```

```
olderror = zeros(maxiter+1,1);

olderror(1) = 0.5*trAtA - trace(C'*(B'*A)) +
    0.5*trace(C'*(B'*B*C)) +
    0.5*trace(B'*diag(newbeta)*B) +
    0.5*trace(C'*(C*diag(newalpha)));

sigma = 1.0e-9; delta = sigma;

for iteration=1:maxiter
  CCt = C*C';
  gradB = B*CCt-A*C'+diag(newbeta)*B;
  Bm  = max(B, (gradB<0)*sigma);
  B   = B - (Bm.*gradB./(Bm*CCt +
        diag(newbeta)*Bm + delta));

  BtB = B'*B;
  gradC = BtB*C-B'*A+C*diag(newalpha);
  Cm  = max(C, (gradC<0)*sigma);
  C   = C - (Cm.*gradC./(BtB*Cm +
        Cm*diag(newalpha) + delta));

  for i = 1:m
    newbeta(i) = gammaB(i)*norm((A(i,:) -
        B(i,:)*C),'fro')^2/(norm(B(i,:),'fro')^2
        + delta);
  end

  for i = 1:n
    newalpha(i) = gammaC(i)*norm((A(:,i) -
        B*C(:,i)),'fro')^2/(norm(C(:,i),'fro')^2
        + delta);
  end

  newerror = 0.5*trAtA - trace(C'*(B'*A)) +
        0.5*trace(C'*(B'*B*C)) +
        0.5*trace(B'*diag(newbeta)*B) +
        0.5*trace(C'*(C*diag(newalpha)));

  errordiff = abs(newerror -
        olderror(iteration));
  olderror(iteration+1) = newerror;
  NablaB = (B*C*C' - A*C' +
        diag(newbeta).*B;
  NablaC = (B'*B*C - B'*A +
        C*diag(newalpha)).*C;

  maxNablaB = max(max(abs(NablaB)));
  maxNablaC = max(max(abs(NablaC)));
  if(maxNablaB < tol && maxNablaC < tol)
    break;
  end

end
resNorm = norm((A-B*C),'fro')^2;
solNorm(1) = norm(B,'fro')^2;
solNorm(2) = norm(C, 'fro')^2;
```